\documentclass[10pt,letterpaper,twocolumn]{article}
\usepackage[T1]{fontenc}
\usepackage[utf8]{inputenc}
\usepackage{lmodern}
\usepackage[margin=0.76in]{geometry}
\usepackage{amsmath,amssymb,booktabs,tabularx,array,graphicx,multirow}
\usepackage{microtype}
\usepackage{url,xurl}
\usepackage{apacite}
\usepackage[font=small,labelfont=bf]{caption}
\usepackage{placeins}
\newcommand{\RUL}{\mathrm{RUL}}

\title{Operating-Condition Normalization in C-MAPSS: Matched Historical Baselines, Multi-Seed Robustness, and Prognostic-Score Trade-offs}
\author{Mohammad Adeel\\[3pt]
\small Department of Electrical Engineering, Prince Sattam bin Abdulaziz University\\
\small Al Kharj, Saudi Arabia\\
\small \texttt{m.adeelsoomro99@gmail.com}\quad
\small ORCID: 0009-0000-9072-494X}
\date{}
\begin{document}
\maketitle

\begin{abstract}
Operating-state changes can obscure turbofan degradation signals, but a reduction in remaining-useful-life (RUL) error after condition-wise standardization may reflect weaknesses of the comparison features instead of a unique benefit of normalization. A controlled retrospective audit examines NASA C-MAPSS FD002 and FD004 with engine-disjoint folds, raw condition identifiers, two constructions of causal history, early-life baselines, and continuous operating-setting correction. Ridge regression, random forests and XGBoost are compared under explicit training-engine-only fitting rules. The previously reported XGBoost FD004 terminal RMSE decreased from 30.38 to 29.08 cycles following six-condition standardization; however, raw sensor values accompanied by condition-matched histories and condition identifiers achieved 28.09 cycles, reversing the naive point-estimate ranking, although the paired-engine interval spans zero. For FD002, normalized XGBoost reduced terminal RMSE yet increased the asymmetric NASA score (12,616 to 13,289). Additional model-seed, condition-count, RUL-cap, grouped-validation and explanation checks test sensitivity rather than postulating a universally optimal representation. The six clusters reproduce known simulator operating regimes; no new condition-discovery algorithm, causal diagnostic method, or field-ready prognostic system is claimed. Since the official test endpoints were also used in earlier exploratory comparisons, all newly examined alternatives are reported as retrospective benchmark evidence, not as results from a pristine confirmatory holdout.
\end{abstract}

\section{Introduction}
RUL prediction estimates remaining operating cycles until a specified end of life. The NASA C-MAPSS simulations provide run-to-failure training sequences and truncated test trajectories for benchmarking \cite{saxena2008,nasa2008}. FD002 and FD004 include six discrete operating conditions that induce large shifts in telemetry independent of degradation. Standardizing sensors within operating regimes is established PHM practice, not a novel learning algorithm \cite{baptista2021}. Yet reported gains depend on sensor histories, the predictor, cap conventions, experimental splits and objective functions.

A previous controlled tabular comparison reported a one-to-two-cycle terminal RMSE reduction when raw sensor values were replaced by six-cluster-normalized values. Its raw temporal history mixed successive measurements from different operating regimes. Such comparisons conflate sensor scale correction with the quality of historical features. Moreover, a method can improve squared error while worsening asymmetric late-prediction penalties. Here we enlarge the comparison to separate these design variables and make negative results visible.

This investigation differs from the extensive factorial replication of \citeA{gupta2026}: the latter reports normalization effects across neural architectures, caps, seeds and windows, whereas the present audit focuses on \emph{matched raw condition labels and within-condition history}, a control absent from the earlier tabular comparison, alongside model-specific NASA-score trade-offs and grouped physical-sensor TreeSHAP. Results are not asserted to exceed recurrent or temporal sequence models; published LSTM/CNN variants often achieve stronger endpoint errors under different evaluation contracts \cite{wang2022,alomari2023}.

\section{Benchmark and experimental protocol}
\subsection{Data, provenance and prospective inputs}
FD002 and FD004 consist of simulated turbofan sensors ($21$ channels), three recorded operating settings, an engine identifier and cycle count. Training data extend to simulated engine failure. Official test data end at varying truncation cycles, and the separate NASA RUL files provide endpoint truth; only that truth enters test scoring. The original FD001/FD003 raw-model results are preserved for historical context but not used to infer multi-condition standardization effects. The multi-condition sample counts are listed in Table~\ref{tab:profile}.
\begin{table*}[!tbp]\centering\small
\caption{NASA C-MAPSS multi-condition subset sizes; test engines with official terminal RUL above the main 125-cycle prediction cap.}\label{tab:profile}
\begin{tabular}{@{}lrrrrr@{}}\toprule
Subset & Train engines & Test engines & Train cycles & Test cycles & RUL $>$125\\\midrule
FD002 & 260 & 259 & 53,759 & 33,991 & 57 \\
FD004 & 249 & 248 & 61,249 & 41,214 & 67 \\
\bottomrule\end{tabular}
\par\vspace{2pt}{\footnotesize Original NASA FD002 and FD004 files; engine IDs are disjoint across official train/test partitions.}
\end{table*}

All transformations in a grouped training fold are estimated only from its training engines. Test and validation operating settings can be used at prediction time, but their sensor distributions never update scalers or regressors. For the official benchmark, the final model and preprocessing are refitted on all designated training engines. The six recorded nominal settings are nearly discrete; six-cluster $k$-means identifies those nominal regimes rather than discovering unobserved physical states. Cluster labels are arbitrary and relabeled 1--6 solely for presentation. Their mapping is tested for stability under multiple random initializations.

\begin{figure*}[!tbp]\centering\includegraphics[width=\linewidth]{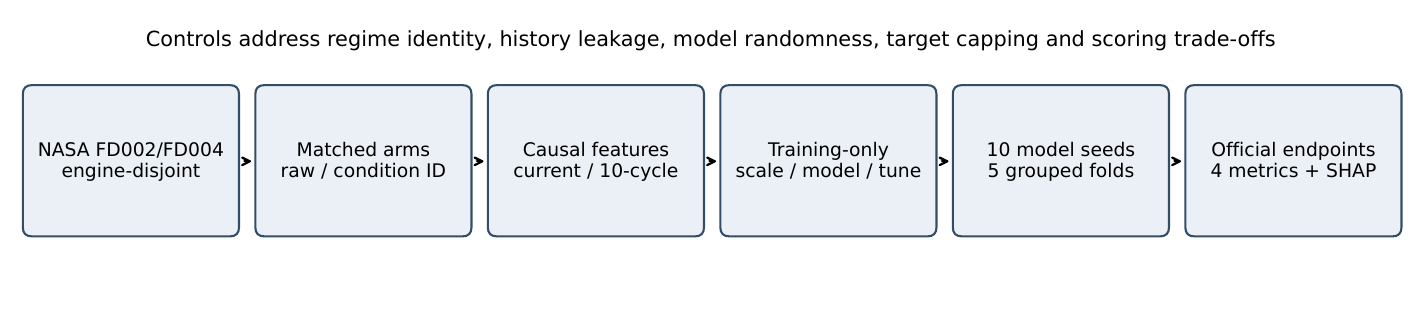}\caption{Matched analysis protocol. The official NASA test endpoint RUL is withheld from fitting and preprocessing. Post-hoc design exploration makes these official test comparisons retrospective rather than confirmatory.}\label{fig:design}\end{figure*}
\begin{figure*}[!tbp]\centering\includegraphics[width=0.90\linewidth]{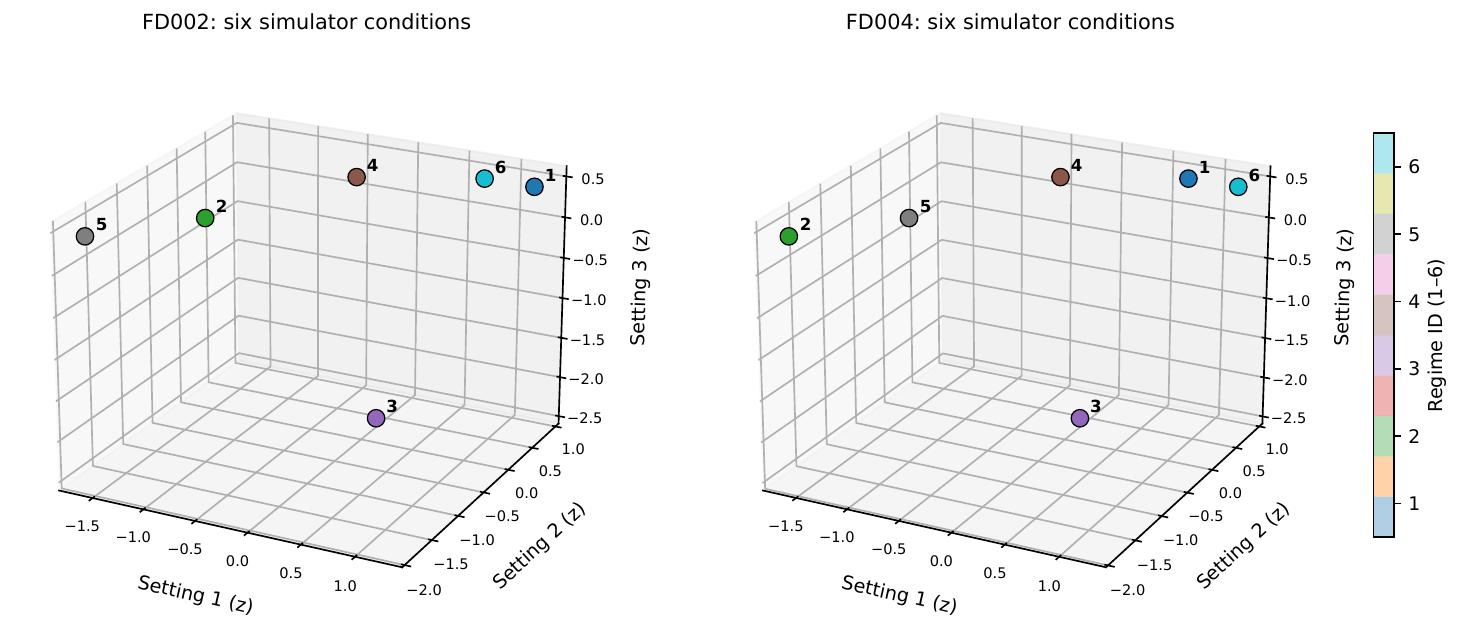}\caption{All three FD002 and FD004 operating settings after training-only scaling, presented in three dimensions. The third setting helps separate nominal configurations that overlap in two-coordinate projections; six clusters correspond to the simulator's operating configurations. Regime indices are 1--6 in the figures and tables; machine-readable CSVs retain zero-based labels.}\label{fig:conditions}\end{figure*}

\subsection{Targets and history conventions}
For training engine $i$ with observed complete life $T_i$, the capped cycle-level target is
\begin{equation}\label{eq:rul}
y_{it}=\min(C,T_i-t),\qquad C=125\text{ cycles}.
\end{equation}
Predictions are truncated to the interval $[0,C]$ for the main model, while official test truths remain uncapped. This intentionally creates a negative error floor for endpoint truths larger than $C$. A sensitivity experiment repeats XGBoost at $C=150$; it is labeled exploratory because cap selection was not reserved exclusively for training-only model selection in the earlier manuscript.

For sensor $s$, the ordinary history feature at time $t$ uses a trailing ten-observation window, with $h_t=\min(10,t)$:
\begin{align}
\bar x_{s,t}&=h_t^{-1}\sum_{j=0}^{h_t-1}x_{s,t-j},\\
\operatorname{sd}_{s,t}&=\left[h_t^{-1}\sum_{j=0}^{h_t-1}(x_{s,t-j}-\bar x_{s,t})^2\right]^{1/2},\\
\text{delta10}_{s,t}&=x_{s,t}-x_{s,t-10}\quad(t>10),
\end{align}
with $\text{delta10}_{s,t}=0$ for $t\le 10$. For early cycles the rolling mean and standard deviation use the shorter available history, with population ($\mathrm{ddof}=0$) standard deviation. A fairer \emph{within-condition history} computes the same statistics from the preceding ten \emph{visits to the same condition} by that engine; successive visits need not be ten consecutive time steps. No value from a later observation is used, and histories never cross engine boundaries.

\subsection{Training-only operating-condition adjustment}
The three settings are standardized using training cycles, then clustered by $k$-means, with $k=6$ and ten initializations. For condition $r$ and sensor $s$, train-cycle means and population standard deviations yield
\begin{equation}\label{eq:reg}
\begin{aligned}
z_{its}&=\frac{x_{its}-\mu_{rs}^{(\mathrm{train})}}{d_{rs}},\\
 d_{rs}&=\begin{cases}
 \sigma_{rs}^{(\mathrm{train})},&\sigma_{rs}^{(\mathrm{train})}\ge10^{-6},\\
 1,&\text{otherwise}.
 \end{cases}
\end{aligned}
\end{equation}
This exactly matches the numerical implementation (no undefined $\epsilon$). Setting clustering and $\mu,\sigma$ are cycle-weighted: longer engines contribute more sensor observations to condition statistics. Regressor fitting, however, gives every engine equal total weight, evaluated separately against unweighted fitting. Training-only early-life normalization estimates condition statistics from cycles 1--30, falling back to the full training condition only when fewer than 50 early observations exist. Continuous correction alternatively subtracts a train-fitted linear regression of each sensor on the standardized settings and standardizes the residuals.

\subsection{Matched representation arms and estimators}
Raw-instantaneous and standardized-instantaneous arms use cycle, three settings and 21 sensor channels. Raw and normalized history arms add each channel's rolling mean, rolling SD and lag difference ($88$ predictors). A \emph{raw+condition-ID} arm retains raw history and adds six one-hot setting-cluster indicators; a \emph{raw+condition-matched-history} arm computes its trailing predictors separately inside each engine-condition stream and also includes the six indicators. Cycle-only ridge and an engine-weighted constant regressor provide weak floors. The six-group lookup is therefore not withheld selectively from the raw comparator.

Ridge regression uses $\alpha=50$ and a train-fitted feature scaler; the random forest uses 65 trees, maximum depth 13, minimum leaf 3, \texttt{max\_features}=0.7, \texttt{max\_samples}=0.8; XGBoost uses 190 trees, depth 4, learning rate 0.07, subsample 0.85, column subsample 0.8, \texttt{min\_child\_weight}=8, $\lambda=2$, histogram trees and squared-error loss. These historical settings were inherited from the preceding exploratory work; the records do not establish that they were selected without examining benchmark results. Thus this is a retrospective audit, not a newly sealed test. A small three-candidate XGBoost search on engine-grouped \emph{training-only} CV is included as a sensitivity test, and its search space is supplied in the code. Ten independent model/condition seeds probe stochastic instability; all train/test engine partitions remain fixed across arms and seeds.

\section{Evaluation and statistics}
\subsection{Four endpoint metrics and NASA asymmetry}
For $n$ test engines, $e_i=\hat y_i-y_i$ is the signed endpoint error. We report
\begin{equation}\label{eq:errors}
\begin{aligned}
\mathrm{RMSE}&=\sqrt{\tfrac1n\sum_i e_i^2},\\
\mathrm{MAE}&=\tfrac1n\sum_i|e_i|,\\
\mathrm{Bias}&=\tfrac1n\sum_i e_i.
\end{aligned}
\end{equation}
where negative bias indicates underprediction, and the dimensionless asymmetric NASA score is
\begin{equation}\label{eq:nasa}
 S=\sum_{i=1}^n\begin{cases}\exp(-e_i/13)-1,&e_i<0,\\\exp(e_i/10)-1,&e_i\ge0.\end{cases}
\end{equation}
Both scores are smaller when better; NASA score additionally penalizes positive and negative errors differently and is strongly affected by error tails. Raw and normalized predictions for the \emph{same engine} are resampled together in a 2000-replicate engine-level percentile bootstrap. These intervals condition on the trained model and the particular test fleet, not on a fresh model fit for each replicate. Six primary RMSE contrasts are treated as a family and explored with Holm adjustment of bootstrap tail proportions, recognizing that bootstrap p-values at the boundary should not be interpreted as definitive null tests.\par

Five engine-grouped folds compare cycle-level capped-target RMSE with identical group membership in both arms. A second, distinct fold-level experiment truncates validation engines at randomly chosen training-domain cutoffs, producing one pseudo-endpoint per engine, with no use of official test labels to select cutoff positions. Grouped folds have overlapping training sets and do not provide five independent datasets. Terminal-style CV and official terminal test RMSE target different estimands, both reported.

\begin{figure*}[!tbp]\centering\includegraphics[width=\linewidth]{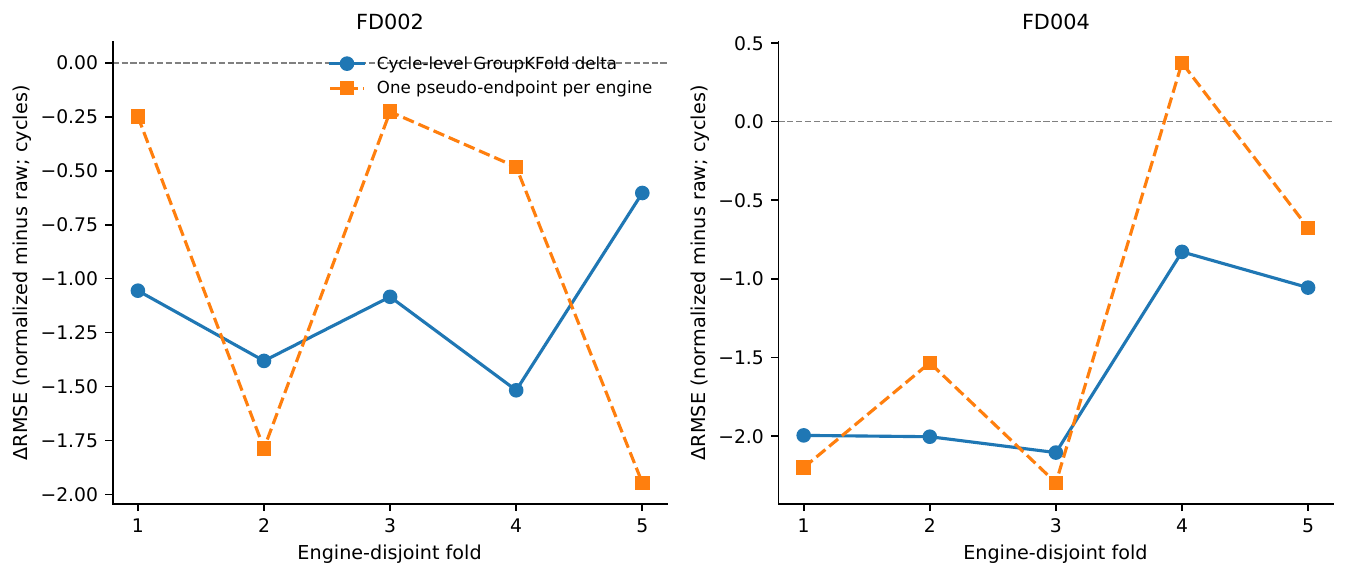}\caption{Paired grouped-fold RMSE differences for XGBoost: cycle-level validation compared with randomly truncated, one-endpoint-per-engine validation. Negative values favor standardization; overlapping fold training sets preclude independent-trial inference.}\label{fig:pairedcv}\end{figure*}

\section{Results}
\subsection{Matched baselines reverse the uncomplicated normalization narrative}
\begin{figure*}[!tbp]\centering\includegraphics[width=\linewidth]{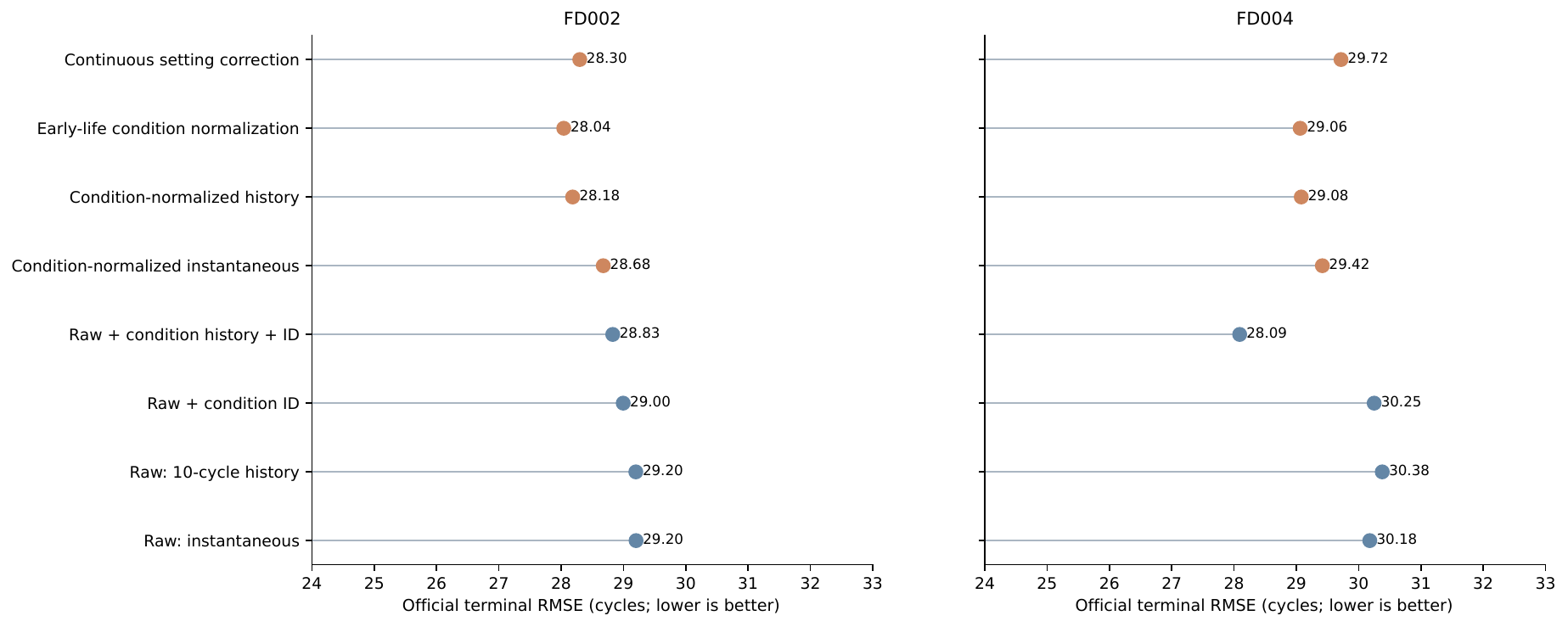}\caption{Official NASA XGBoost test RMSE (lower is better) across eight matched representation arms. The raw condition-matched-history representation addresses the originally handicapped comparator. This ablation is retrospective and does not constitute a new independent test.}\label{fig:ablation}\end{figure*}

Table~\ref{tab:arms} and Fig.~\ref{fig:ablation} compare the exact official endpoint test predictions under multiple representations. On FD004, the raw condition-matched-history arm achieves lower RMSE than the normalized-history arm. The normalized-minus-fair-raw FD004 RMSE contrast is $+0.99$ cycles with a paired engine-bootstrap 95\% interval of $[-0.24,2.19]$ cycles. It reverses the point-estimate ranking but does not demonstrate a statistically decisive difference. This shows that historic raw trailing statistics were disadvantaged by mixing regime changes. Raw feature values can remain useful when condition labels and causal histories are both correctly supplied. On FD002, early-life condition baselining also performs at least as well as pooled six-condition normalization in this fixed configuration.
\begin{table*}[!tbp]\centering\small
\caption{Retrospective official-test RMSE (cycles) of matched representation arms; smaller values are better. Fixed historic model configurations; not tuned against test truth.}\label{tab:arms}
\begin{tabular}{@{}llrr@{}}\toprule
Arm & Estimator & FD002 & FD004\\\midrule
Raw instantaneous & XGBoost & 29.20 & 30.18 \\
Raw + ordinary history & XGBoost & 29.20 & 30.38 \\
Raw + condition ID & XGBoost & 29.00 & 30.25 \\
Raw + condition ID + matched history & XGBoost & 28.83 & 28.09 \\
Normalized instantaneous & XGBoost & 28.68 & 29.42 \\
Normalized + ordinary history & XGBoost & 28.18 & 29.08 \\
Early-life condition normalization & XGBoost & 28.04 & 29.06 \\
Continuous setting correction & XGBoost & 28.30 & 29.72 \\
Cycle-only ridge & Ridge & 40.92 & 50.86 \\
Engine-weighted constant & Constant & 53.92 & 54.65 \\
\bottomrule\end{tabular}
\par\vspace{2pt}{\footnotesize All preprocessing statistics are learned on the original training engines; condition IDs for raw arms are training-fitted settings clusters.}
\end{table*}

\subsection{Balanced endpoint metrics and the score counterexample}
Table~\ref{tab:metrics} reports all four metrics for both raw and regime-adjusted arms and all three model families. Although fixed-seed XGBoost FD002 RMSE falls by about one cycle, its asymmetric NASA score increases, exposing a consequential trade-off hidden by RMSE-only reporting. The same score worsened under normalization for eight of ten FD002 XGBoost model seeds, despite the RMSE difference being negative for every one of those seeds. The FD002 score reversal comes from the underprediction tail: its score contribution rises from approximately 11,924 to 12,602 after normalization, whereas the overprediction contribution is nearly unchanged (693 versus 687). Thus average signed bias alone conceals changes in the exponential penalty tails. The upper endpoint of the FD002 XGBoost paired RMSE interval lies close to zero; its sign should not be treated as decisive under the six-contrast multiplicity problem.\par
\begin{table*}[!tbp]\centering\small
\caption{Raw versus six-condition-normalized official-test endpoint metrics for the same engines. RMSE, MAE and signed bias are in cycles; NASA score is dimensionless and smaller is better.}\label{tab:metrics}
\begin{tabular}{@{}lllrrrr@{}}\toprule
Subset & Model & Representation & RMSE & MAE & Bias & NASA score\\\midrule
FD002 & Ridge & Raw & 31.86 & 23.78 & -10.07 & 16,965\\
 & & Normalized & 31.01 & 23.09 & -9.67 & 14,854\\
\addlinespace[2pt]
FD002 & RF & Raw & 28.79 & 20.30 & -8.64 & 11,955\\
 & & Normalized & 28.01 & 19.39 & -8.71 & 11,618\\
\addlinespace[2pt]
FD002 & XGBoost & Raw & 29.20 & 20.64 & -9.95 & 12,616\\
 & & Normalized & 28.18 & 19.38 & -9.60 & 13,289\\
\addlinespace[2pt]
FD004 & Ridge & Raw & 35.32 & 28.21 & -8.91 & 15,403\\
 & & Normalized & 33.04 & 25.77 & -9.33 & 10,885\\
\addlinespace[2pt]
FD004 & RF & Raw & 29.91 & 22.29 & -8.23 & 6,856\\
 & & Normalized & 28.66 & 21.22 & -8.65 & 5,772\\
\addlinespace[2pt]
FD004 & XGBoost & Raw & 30.38 & 22.58 & -8.86 & 7,462\\
 & & Normalized & 29.08 & 20.99 & -9.20 & 6,538\\
\addlinespace[2pt]
\bottomrule\end{tabular}
\par\vspace{2pt}{\footnotesize Bias is predicted minus observed RUL; negative indicates underprediction. FD002 XGBoost NASA score worsens under normalization.}
\end{table*}

\begin{table*}[!tbp]\centering\small
\caption{Paired test-engine 95\% bootstrap intervals of normalized-minus-raw differences, 2000 resamples per comparison. Six RMSE contrasts receive an exploratory Holm adjustment; these conditional intervals do not account for model selection or reused test endpoints.}\label{tab:ci}
\begin{tabular}{@{}lllrrr@{}}\toprule
Subset & Model & Metric & $\Delta$ & 95\% CI & Holm $p$\\\midrule
FD002 & Ridge & RMSE & -0.85 & [-1.27, -0.42] & 0.000\\
FD002 & Ridge & MAE & -0.68 & [-1.24, -0.16] & --\\
FD002 & Ridge & Bias & +0.40 & [-0.16, +0.93] & --\\
FD002 & Ridge & NASAScore & -2,111 & [-3,845, -626] & --\\
FD002 & RF & RMSE & -0.77 & [-1.56, +0.00] & 0.080\\
FD002 & RF & MAE & -0.91 & [-1.69, -0.10] & --\\
FD002 & RF & Bias & -0.07 & [-1.02, +0.86] & --\\
FD002 & RF & NASAScore & -337 & [-3,147, 2,382] & --\\
FD002 & XGBoost & RMSE & -1.01 & [-2.09, -0.04] & 0.080\\
FD002 & XGBoost & MAE & -1.26 & [-2.13, -0.41] & --\\
FD002 & XGBoost & Bias & +0.35 & [-0.67, +1.43] & --\\
FD002 & XGBoost & NASAScore & 673 & [-2,119, 3,706] & --\\
FD004 & Ridge & RMSE & -2.28 & [-3.00, -1.55] & 0.000\\
FD004 & Ridge & MAE & -2.44 & [-3.21, -1.65] & --\\
FD004 & Ridge & Bias & -0.42 & [-1.33, +0.49] & --\\
FD004 & Ridge & NASAScore & -4,518 & [-7,873, -1,820] & --\\
FD004 & RF & RMSE & -1.24 & [-2.17, -0.32] & 0.052\\
FD004 & RF & MAE & -1.07 & [-2.03, -0.11] & --\\
FD004 & RF & Bias & -0.42 & [-1.49, +0.62] & --\\
FD004 & RF & NASAScore & -1,084 & [-1,980, -203] & --\\
FD004 & XGBoost & RMSE & -1.30 & [-2.39, -0.27] & 0.052\\
FD004 & XGBoost & MAE & -1.59 & [-2.68, -0.53] & --\\
FD004 & XGBoost & Bias & -0.33 & [-1.44, +0.78] & --\\
FD004 & XGBoost & NASAScore & -924 & [-2,218, 367] & --\\
\bottomrule\end{tabular}
\end{table*}

\begin{figure*}[!tbp]\centering\includegraphics[width=\linewidth]{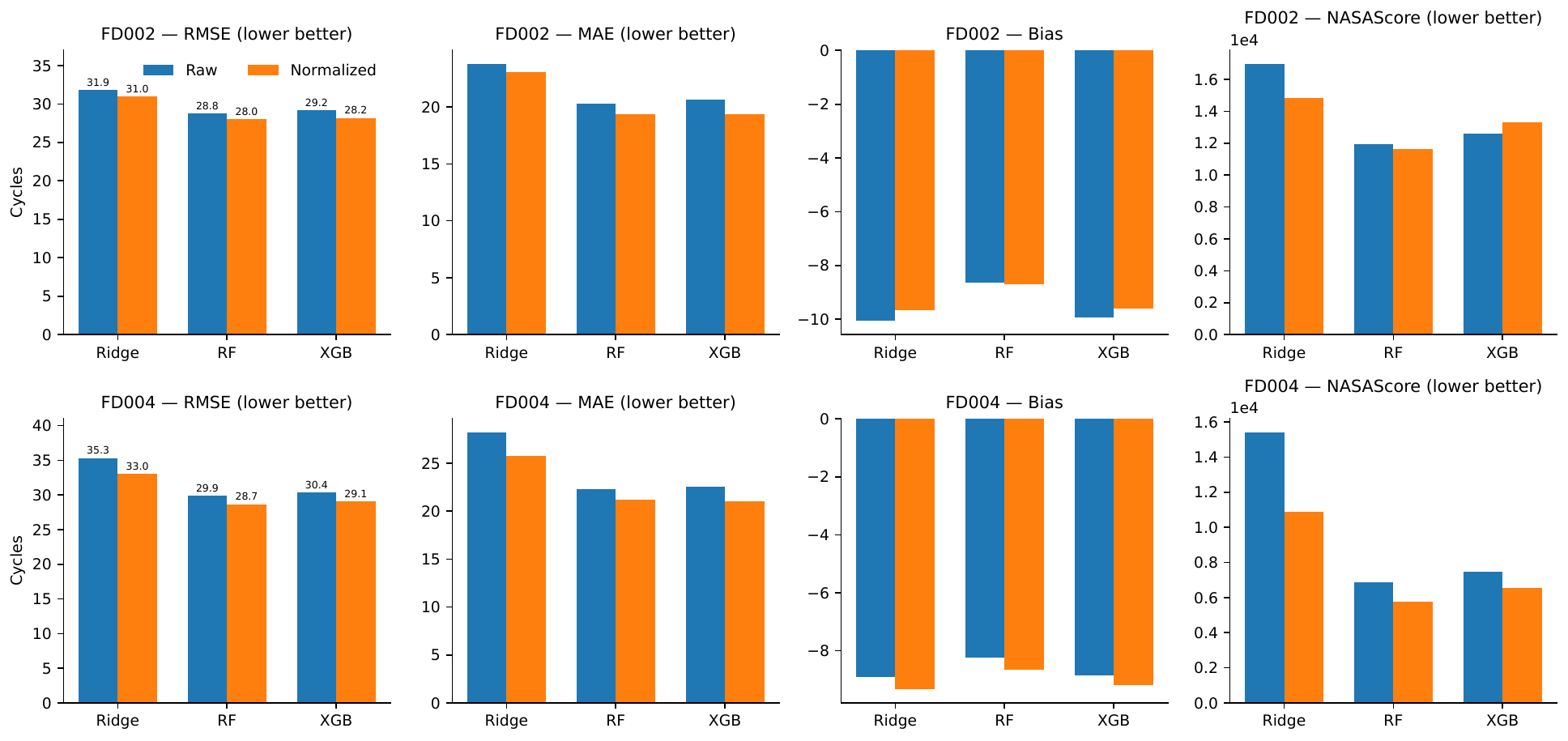}\caption{Raw versus normalized NASA endpoint RMSE, MAE, signed bias and dimensionless NASA score; lower is better except that bias is interpreted relative to zero. The FD002 XGBoost score reverses the RMSE ordering.}\label{fig:metrics}\end{figure*}

\subsection{Seed dispersion, group folds and sensitivity tests}
\begin{figure*}[!tbp]\centering\includegraphics[width=\linewidth]{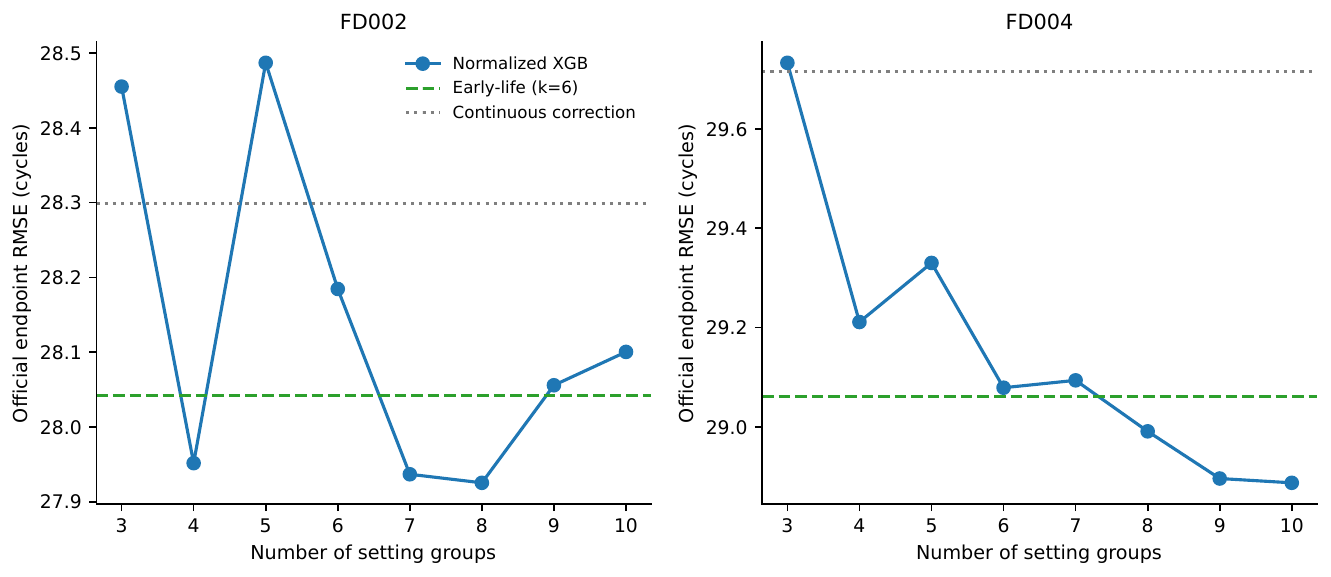}\caption{XGBoost endpoint sensitivity to setting-cluster count $k$ from 3 to 10. Early-life condition baselining and continuous setting regression provide alternative, non-exclusive normalizers; the curves are descriptive retrospective comparisons.}\label{fig:k}\end{figure*}

Ten matched seeds allow model stochasticity to be separated from fleet-resampling uncertainty (Table~\ref{tab:seed}). Model-seed differences do not replace datasets or provide a fresh official holdout. The same table also summarizes the five paired CV fold differences in the original cycle-level task; the pseudo-endpoint fold analysis is visualized in Fig.~\ref{fig:pairedcv}. Results across $k=3,\ldots,10$ (Fig.~\ref{fig:k}) and the 150-cycle cap sensitivity are described in Table~\ref{tab:sens}.\par
\begin{table*}[!tbp]\centering\small
\caption{Matched normalized-minus-raw $\Delta\text{RMSE}$ (cycles): ten paired model seeds on the official test fleet and five engine-grouped cycle-level CV folds. Negative values favor normalization.}\label{tab:seed}
\begin{tabular}{@{}llrrrr@{}}\toprule
Subset & Model & Seeds & Mean $\pm$ SD & Fold mean & Fold range\\\midrule
FD002 & Ridge & 10 & -0.85 $\pm$ 0.00 & -0.44 & [-0.79, -0.19] \\
FD002 & RF & 10 & -0.73 $\pm$ 0.11 & -0.62 & [-1.03, -0.13] \\
FD002 & XGBoost & 10 & -1.16 $\pm$ 0.12 & -1.13 & [-1.52, -0.60] \\
FD004 & Ridge & 10 & -2.28 $\pm$ 0.00 & -1.48 & [-2.03, -0.78] \\
FD004 & RF & 10 & -1.27 $\pm$ 0.11 & -0.77 & [-1.30, -0.15] \\
FD004 & XGBoost & 10 & -1.41 $\pm$ 0.17 & -1.60 & [-2.11, -0.83] \\
\bottomrule\end{tabular}
\par\vspace{2pt}{\footnotesize The CV training partitions overlap; the seed spread is conditional on this official test fleet. These are distinct uncertainty sources.}
\end{table*}

\begin{figure*}[!tbp]\centering\includegraphics[width=\linewidth]{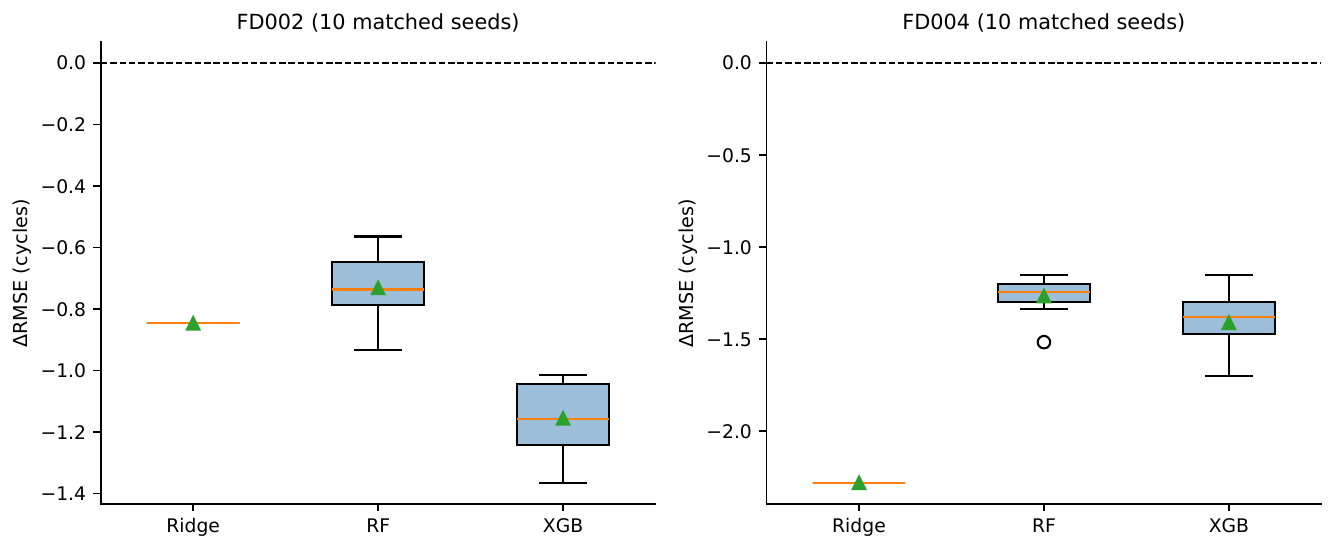}\caption{Ten paired model and clustering seeds per subset. The paired distribution is a stochastic-fit sensitivity conditional on one official test fleet; the zero line marks no normalized-minus-raw RMSE effect.}\label{fig:seeds}\end{figure*}
\begin{table*}[!tbp]\centering\small
\caption{Retrospective XGBoost endpoint RMSE sensitivity (cycles). For rows listing two arms, the third and fourth columns report raw and normalized RMSE respectively.}\label{tab:sens}
\begin{tabular}{@{}llrr@{}}\toprule
Subset & Variant & RMSE A & RMSE B\\\midrule
FD002 & $k=6$ baseline & 28.18 & --\\
FD002 & $k=8$ (retrospective best) & 27.93 & --\\
FD002 & Cap=150, raw / normalized & 27.24 & 26.06\\
FD002 & Cycle-weighted raw / normalized & 28.31 & 27.38\\
FD002 & CV-selected raw / normalized & 28.78 & 27.96\\
FD004 & $k=6$ baseline & 29.08 & --\\
FD004 & $k=10$ (retrospective best) & 28.89 & --\\
FD004 & Cap=150, raw / normalized & 27.29 & 26.25\\
FD004 & Cycle-weighted raw / normalized & 30.01 & 28.54\\
FD004 & CV-selected raw / normalized & 29.68 & 28.71\\
\bottomrule\end{tabular}
\par\vspace{2pt}{\footnotesize “Best $k$” is descriptive and not a valid test-selected hyperparameter. Tuned alternatives use only the training-domain threefold CV for candidate selection; official test evaluation remains retrospective.}
\end{table*}

A small three-candidate XGBoost hyperparameter search using only engine-grouped training folds selected depth 5 and 220 boosting rounds in both arms. The resulting retrospective terminal RMSE is 28.78 raw versus 27.96 normalized on FD002 and 29.68 versus 28.71 on FD004. Importantly, the FD002 score trade-off remains: the NASA score increases from 11,790 to 13,067 after normalization. This does not establish that the selected parameters generalize better to an unseen cohort.\par

\subsection{Capping and the endpoint estimand}
\begin{figure*}[!tbp]\centering\includegraphics[width=\linewidth]{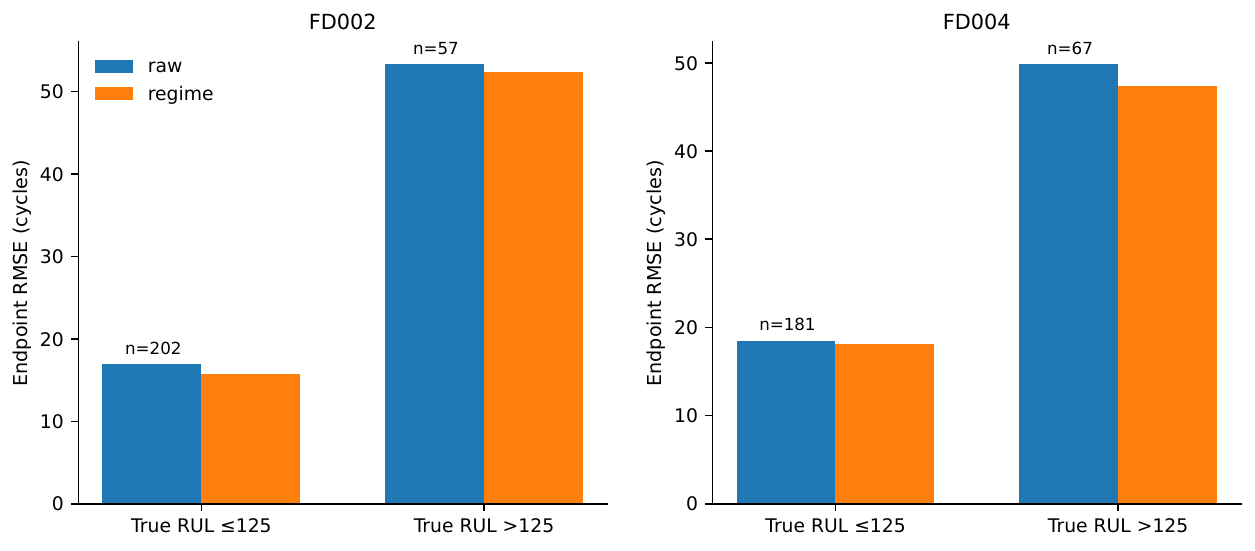}\caption{XGBoost terminal RMSE separated at true RUL = 125, with engine counts annotated. Raw and normalized model predictions share the same 125-cycle ceiling; all true test targets are uncapped.}\label{fig:cap}\end{figure*}

RUL above the 125-cycle prediction cap cannot be recovered by the capped predictor. In the official test fleets, 57/259 FD002 and 67/248 FD004 engines have target RUL larger than 125. Thus the full-test RMSE includes a systematic lower-tail bias for high-RUL engines; the subset with $\RUL\le125$ is a distinct, less cap-constrained estimand (Fig.~\ref{fig:cap}). Raising the training and output cap to 150 changes the target distribution as well as the output ceiling, and cannot be interpreted as pure preprocessing improvement.

\subsection{Aggregating and stress-testing explanations}
\begin{figure*}[!tbp]\centering\includegraphics[width=\linewidth]{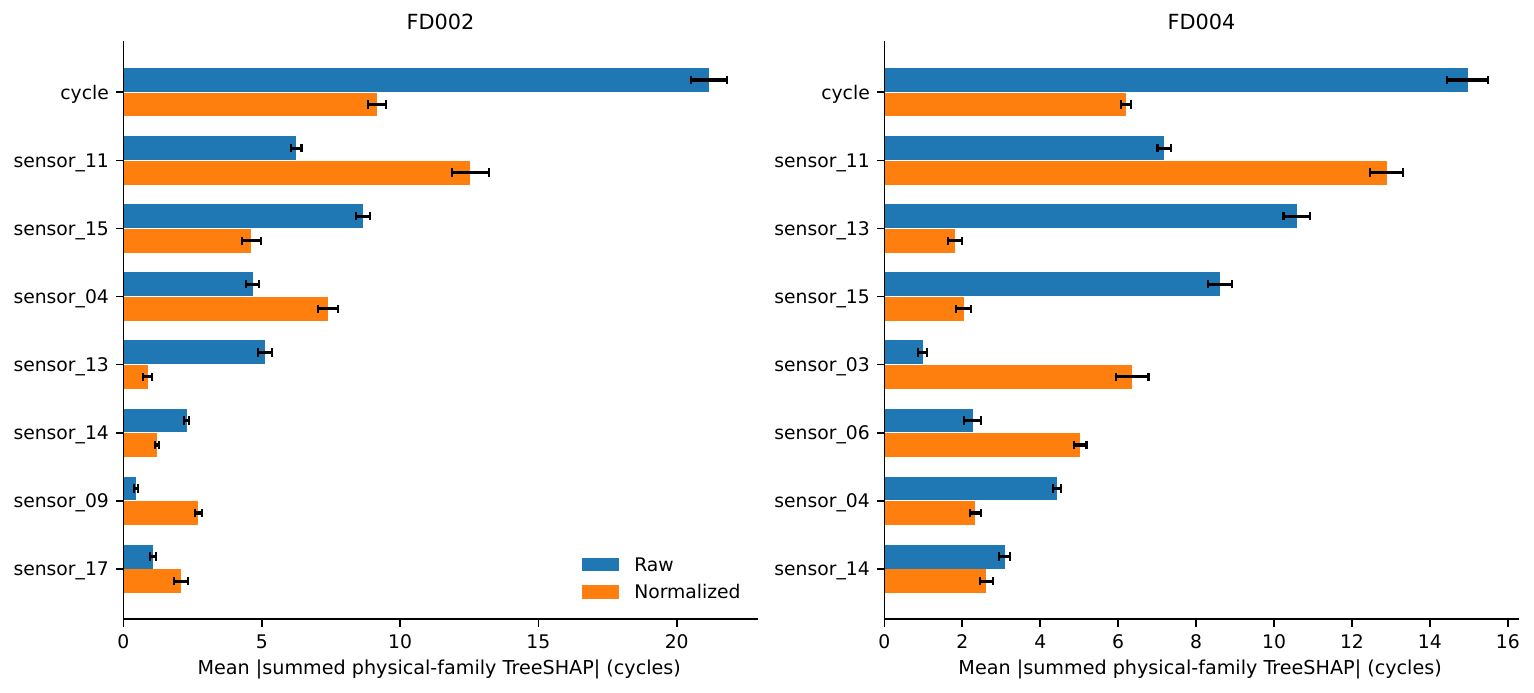}\caption{Model-seed dispersion of TreeSHAP importance grouped by physical sensor family; bars show means and whiskers show across-seed standard deviations. Correlation and feature-coordinate changes prevent causal interpretation.}\label{fig:shap}\end{figure*}

TreeSHAP gives additive contributions $\phi_f$ on the fitted model's RUL-output scale; the sum of contributions reproduces the model's prediction, not the simulator's physical failure mechanism \cite{lundberg2017,lundberg2020}. Sensor histories are highly correlated. For physical sensor $s$, we \emph{first sum signed SHAP contributions} from raw, mean, SD and difference variants and only then take absolute values:
\begin{equation}\label{eq:shapfamily}
\begin{aligned}
\Phi_s(i) &= \sum_{f\in\mathcal{F}_s}\phi_f(i),\qquad
\mathcal{F}_s = \{s, s_{\mathrm{ma10}}, s_{\mathrm{sd10}}, s_{\mathrm{delta10}}\},\\
I_s &= \frac{1}{n}\sum_{i=1}^{n}\lvert\Phi_s(i)\rvert.
\end{aligned}
\end{equation}
This retains grouped additivity and avoids interpreting attribution transfer between near-duplicates as proof of new physics. Results are compared across seeds on both training-engine samples and official test endpoints, with uncertainty coming from seed dispersion; endpoint bootstrap across engines is a separate conditional uncertainty source. The dominant families and the cycle contribution appear in Table~\ref{tab:shap} and Fig.~\ref{fig:shap}.\par
\begin{table*}[!tbp]\centering\small
\caption{Physical-family XGBoost TreeSHAP at official test endpoints. Each physical sensor is grouped using signed sums across instantaneous, rolling mean, rolling SD and lag difference, followed by absolute value; variability is across ten fitted seeds.}\label{tab:shap}
\begin{tabular}{@{}llrrr@{}}\toprule
Subset & Family & Raw $\pm$ SD & Norm. $\pm$ SD & Norm. CI\\\midrule
FD002 & cycle & 21.18 $\pm$ 0.65 & 9.16 $\pm$ 0.32 & $\left[8.37,\,10.05\right]$\\
FD002 & sensor\_11 & 6.25 $\pm$ 0.19 & 12.54 $\pm$ 0.67 & $\left[11.13,\,14.27\right]$\\
FD002 & sensor\_15 & 8.67 $\pm$ 0.25 & 4.63 $\pm$ 0.34 & $\left[3.85,\,5.41\right]$\\
FD002 & sensor\_04 & 4.68 $\pm$ 0.24 & 7.38 $\pm$ 0.36 & $\left[6.63,\,8.41\right]$\\
FD002 & sensor\_13 & 5.11 $\pm$ 0.25 & 0.88 $\pm$ 0.16 & $\left[0.61,\,1.14\right]$\\
FD004 & cycle & 14.97 $\pm$ 0.52 & 6.20 $\pm$ 0.12 & $\left[5.73,\,6.63\right]$\\
FD004 & sensor\_11 & 7.18 $\pm$ 0.17 & 12.89 $\pm$ 0.42 & $\left[11.38,\,14.20\right]$\\
FD004 & sensor\_13 & 10.58 $\pm$ 0.34 & 1.81 $\pm$ 0.18 & $\left[1.47,\,2.30\right]$\\
FD004 & sensor\_15 & 8.61 $\pm$ 0.30 & 2.03 $\pm$ 0.19 & $\left[1.79,\,2.53\right]$\\
FD004 & sensor\_03 & 0.98 $\pm$ 0.10 & 6.36 $\pm$ 0.42 & $\left[5.27,\,7.18\right]$\\
\bottomrule\end{tabular}
\par\vspace{2pt}{\footnotesize A ranking shift does not establish fault causality or independent explanation stability; training-sample family attributions are exported separately.}
\end{table*}

\begin{figure*}[!tbp]\centering\includegraphics[width=\linewidth]{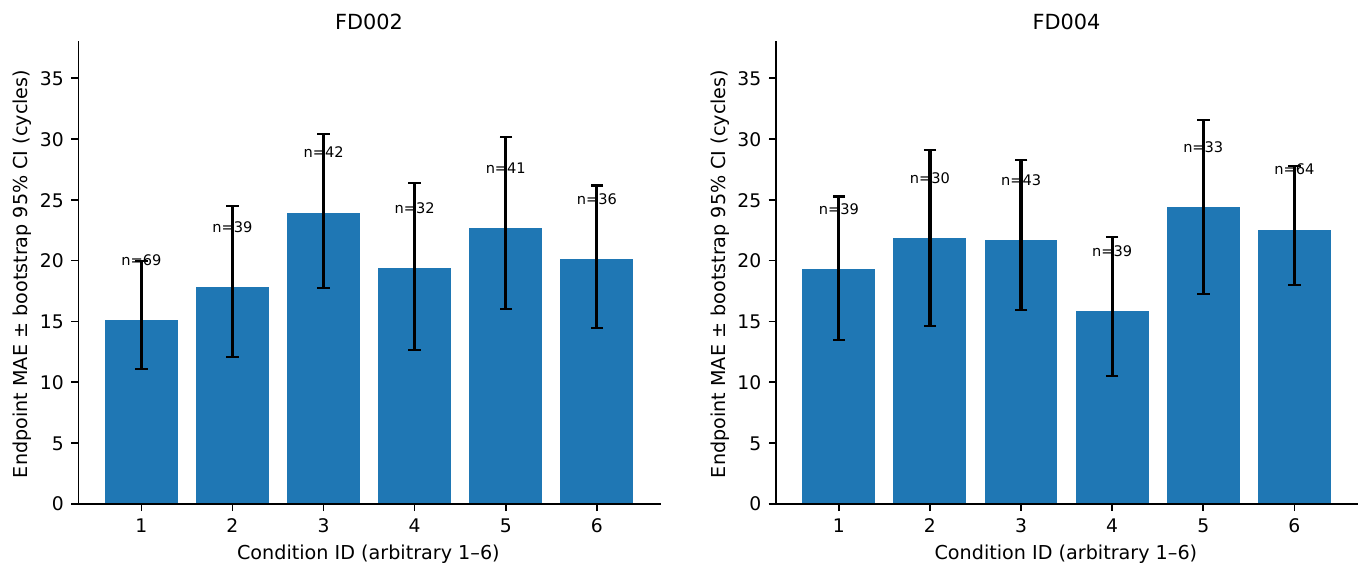}\caption{Normalized XGBoost official test endpoint MAE per operating condition, with engine sample size $n$ and 95\% engine-bootstrap intervals. Condition indexing is arbitrary (1 to 6 in each dataset).}\label{fig:cond}\end{figure*}

Condition-wise errors have small and unequal sample sizes; Fig.~\ref{fig:cond} reports counts and paired-independent within-condition engine-bootstrap intervals rather than interpreting raw point-estimate rankings as meaningful differences. Early-life and continuous controls, seed variation, and matched raw history place clear limits on interpreting any single SHAP ranking as robust physical prognostic knowledge.

\section{Discussion}
\subsection{Confounding, robust baselines and comparison with prior work}
The principal insight is methodological rather than architectural: operating-regime normalization cannot be credited with every improvement obtained when the competing raw model uses a less informative historical representation. The fair condition-specific-history baseline is especially competitive in FD004. Continuous settings regression, early-life statistics and multiple $k$ values are reasonable established alternatives, so six-cluster standardization should be understood as one design choice. This accords with the broader baselining literature \cite{baptista2021} and complements, rather than supersedes, the much larger neural architecture and normalization factorial study of \citeA{gupta2026}.

Published feature-engineered models and deep temporal approaches report lower errors under their own evaluation protocols. For example, the AFICv feature-selection study of \citeA{alomari2023} reports approximately 23.0-cycle FD002 and 22.3-cycle FD004 RMSE, substantially below the 28--29-cycle tabular results here; \citeA{wang2022} provides a further recurrent/temporal comparison. These are contextual figures, \emph{not} a protocol-matched benchmark, because target capping, feature engineering and model selection differ. This paper does not claim a new state of the art. Its useful contribution is a negative control and reproducible accounting of trade-offs: a small improvement in RMSE can coincide with a worse NASA score; improved within-condition raw histories can remove or reverse the normalized advantage.

\subsection{Limitations and selection bias}
The official test data had already been scored for the original model, temporal-history ablation and standardization comparison before the present revision. Thus experiments on additional arms are an \emph{exploratory reuse} of a benchmark fleet, and their differences are susceptible to test-set adaptivity. Bootstrap intervals only resample the fixed fleet and do not correct model-selection or multiple-comparison bias. Model-seed spread and paired folds quantify other dimensions but do not create independent external validation. Settings identify six mostly discrete simulator conditions; condition recovery is not a new discovery algorithm. The pooled training condition mean mixes healthy and degraded cycles; early-life baselines partially address this, without access to true individual health states.

Neither TreeSHAP additivity nor a shifted attribution distribution validates a fault mechanism. Without engine component wear states or causal interventions, grouped physical-sensor attributions remain descriptive. The experiment uses synthetic C-MAPSS trajectories, not instrumented real aircraft data. Any operational maintenance or safety claim would require prospective field evidence.

\section{Conclusions}
A controlled audit of C-MAPSS FD002 and FD004 demonstrates that six-condition normalization is not guaranteed to outperform a raw comparator when both receive appropriate condition information and causal histories. The FD004 matched raw condition-history XGBoost model attains 28.09-cycle RMSE versus 29.08 for the normalized history model; FD002 normalized XGBoost reduces RMSE relative to the original raw pipeline while worsening the asymmetric NASA score. Ten-seed, fold-paired, condition-count, capping, weighting and grouped attribution checks expose sensitivity rather than establishing universal superiority. The key implication is that leakage-safe preprocessing comparisons also require structurally fair feature construction and balanced, explicitly defined evaluation metrics.

\section*{Data and code availability}
The NASA C-MAPSS benchmark is available from the NASA Prognostics Center of Excellence (\cite{nasa2008}); an archived dataset copy is available at \url{https://doi.org/10.5281/zenodo.15346912}. The accompanying source archive provides analysis code, NASA records, seed-level predictions, fold outputs, explanation summaries, figures and an executable notebook. No public code-repository DOI is claimed until a release has actually been deposited.

\section*{Acknowledgements}
NASA's Prognostics Center of Excellence is acknowledged for releasing the simulated turbofan degradation benchmark.
\section*{Funding}
The author received no specific funding for this work.
\section*{Conflict of interest}
The author has no conflict of interest to declare.

\bibliographystyle{apacite}
\bibliography{references}
\end{document}